\documentclass[a4paper,11pt,twocolumn,twoside]{article}
\usepackage{graphicx}
\usepackage{sepln_en}
\usepackage{fullname_esp}
\usepackage[utf8]{inputenc}
\usepackage[english]{babel}
\usepackage{url}
\input epsf
\usepackage{subcaption}
\usepackage{xcolor}

\title{Automating Early Disease Prediction Via Structured and Unstructured Clinical Data}

\author {\textbf{Ane G. Domingo-Aldama$^1$, Marcos Merino Prado$^1$, Alain García Olea$^2$,}\\ 
\textbf{Josu Goikoetxea$^1$, Koldo Gojenola$^1$, Aitziber Atutxa$^1$}\\
$^1$University of the Basque Country (EHU), $^2$Basurto University Hospital\\
ane.garciad@ehu.eus, mmerinoprado@gmail.com, alain.garciaolea@osakidetza.eus,\\
josu.goikoetxea@ehu.eus, koldo.gojenola@ehu.eus, aitziber.atucha@ehu.eus\\
}

\seplntranstitle{Automatización de la predicción temprana de enfermedades mediante datos clínicos estructurados y no estructurados}

\seplnclave{Predicción Temprana, Progresión de la Fibrilación Auricular, Historias Clínicas Electrónicas, Procesamiento del Lenguaje Natural.}

\seplnresumen{Este estudio presenta una metodología totalmente automatizada para estudios de predicción temprana en entornos clínicos, aprovechando la información extraída de informes de alta hospitalaria no estructurados. El proceso propuesto utiliza los informes de alta para respaldar los tres pasos principales de la predicción temprana: selección de cohortes, generación de conjuntos de datos y etiquetado de resultados. Mediante el procesamiento de los informes de alta con técnicas de procesamiento del lenguaje natural, podemos identificar de manera eficiente las cohortes de pacientes relevantes, enriquecer los conjuntos de datos estructurados con variables clínicas adicionales y generar etiquetas de alta calidad sin intervención manual. Este enfoque aborda el principal problema de los registros médicos electrónicos (RME) codificados que son los datos faltantes o incompletos, capturando información clínicamente relevante que a menudo está infrarrepresentada. Evaluamos la metodología en el contexto de la predicción de la progresión de la fibrilación auricular (FA), demostrando que los modelos predictivos entrenados con conjuntos de datos enriquecidos con información de informes de alta logran una mayor precisión y correlación con los resultados reales en comparación con los modelos entrenados únicamente con datos estructurados de RME, al tiempo que superan las puntuaciones clínicas tradicionales. Estos resultados demuestran que la automatización de la integración de texto clínico no estructurado puede agilizar los estudios de predicción temprana, mejorar la calidad de los datos y aumentar la fiabilidad de los modelos predictivos para la toma de decisiones clínicas.}

\seplnkey{Early Prediction, Atrial Fibrillation Progression, Electronic Health Records, Natural Language Processing.}

\seplnabstract{This study presents a fully automated methodology for early prediction studies in clinical settings, leveraging information extracted from unstructured discharge reports. The proposed pipeline uses discharge reports to support the three main steps of early prediction: cohort selection, dataset generation, and outcome labeling. By processing discharge reports with natural language processing techniques, we can efficiently identify relevant patient cohorts, enrich structured datasets with additional clinical variables, and generate high-quality labels without manual intervention. This approach addresses the frequent issue of missing or incomplete data in codified electronic health records (EHR), capturing clinically relevant information that is often underrepresented. We evaluate the methodology in the context of predicting atrial fibrillation (AF) progression, showing that predictive models trained on datasets enriched with discharge report information achieve higher accuracy and correlation with true outcomes compared to models trained solely on structured EHR data, while also surpassing traditional clinical scores. These results demonstrate that automating the integration of unstructured clinical text can streamline early prediction studies, improve data quality, and enhance the reliability of predictive models for clinical decision-making.}

\firstpageno{1}

\begin{document}

% la siguiente instrucción sólo se debe usar si el abstract sobrescribe el texto
% la longitud variará según se necesite

%\setlength\titlebox{20cm} % se aumenta el tamaño del espacio reservado para datos de título

\label{firstpage} \maketitle

\clearpage

\section{Introduction}
Early diagnosis (ED) has garnered significant attention in both Artificial Intelligence (AI) and medicine due to its transformative potential in personalized medicine. By identifying a condition at its initial stages, clinicians can intervene sooner, prevent complications, and tailor treatments to each patient which leads to improved outcomes and reduced healthcare costs. The success of these ED models heavily depends on the diversity, volume, and granularity of the data available \cite{ng2016early,alzu2021electronic}. 

Modern AI systems for personalized and predictive medicine predominantly employ machine learning (ML) models that rely exclusively on structured, tabular data (including: demographic information, laboratory measurements, coded diagnoses, etc.). This reliance has led to an increasing dependence on structured electronic health records (EHRs) \cite{ristevski2018big}. EHRs, now central to healthcare systems worldwide, provide a rich repository of both structured and unstructured patient information, playing a crucial role in clinical decision-making \cite{holmes2021electronic,alzu2021electronic}.

Despite the rich information contained in structured EHRs, they present notable challenges, primarily stemming from the manual annotation and conversion of medical data into standardized structured and discrete fields. This process is prone to errors, missing values, and inconsistencies, which can significantly impact data reliability and the overall quality of the predictive models built upon this information \cite{garcia2021role,botsis2010secondary,alzu2021electronic}. 

Efforts to standardize medical coding with systems like ICD-10, OPCS, and SNOMED have mitigated some challenges. However, no universal guidelines exist regarding the level of detail required for clinical documentation. Consequently, not all information relevant to specific prediction tasks is captured in structured EHRs, often necessitating extra manual annotation (a time-consuming process that introduces variability and potential human error).

%\footnote{\url{https://icd.who.int/en/}}
%\footnote{\url{https://classbrowser.nhs.uk/ref_books/OPCS-4.10_NCCS-2024.pdf}}
%\footnote{\url{https://www.snomed.org/}}

Furthermore, ED studies require a cohort selection process to identify patients suitable for the task. This step typically depends on structured EHR data or, when key information is missing, on manual review of clinical records. As a result, the process becomes labor intensive, slows down research workflows, and introduces variability and potential human error \cite{chen2025enhancing,jin2024matching}.

In this context, the present study introduces a methodology that automates the key steps required for early prediction tasks, including cohort selection, dataset generation, and patient labeling. The approach combines structured tabular data derived from EHRs with semi-structured discharge reports in free-text format, processed through a natural language processing (NLP) pipeline. This methodology reduces the need for manual annotation while simultaneously improving the quality and completeness of structured tabular data by leveraging information extracted from discharge reports to mitigate missing values or codification errors. The primary contribution of this work lies in the design of this end-to-end methodology and data generation pipeline, rather than in proposing a novel predictive modeling architecture.

\begin{figure}[htb]
    \centering
    \includegraphics[width=.5\textwidth]{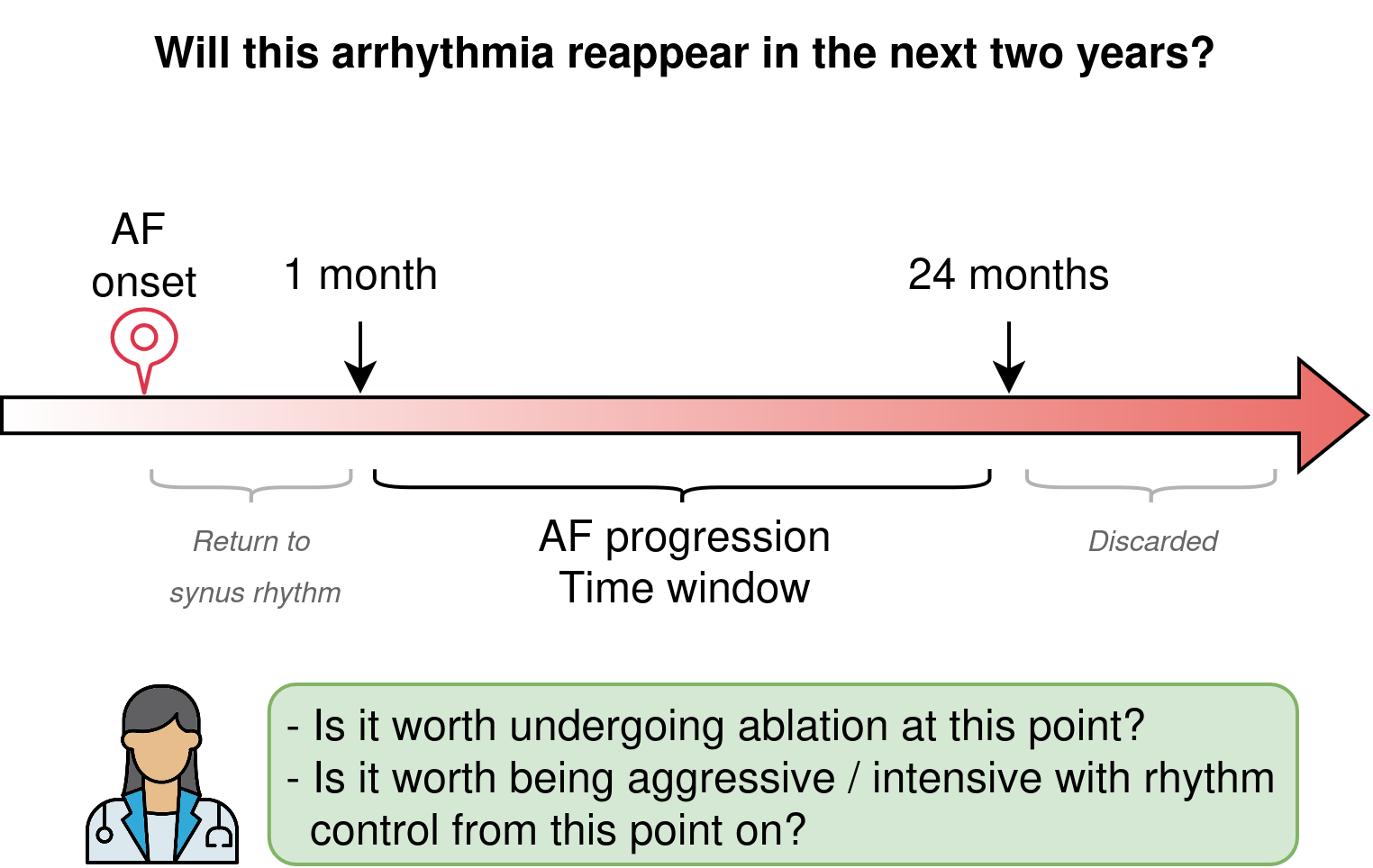}
    \caption{\textbf{Clinical scenario of AF progression.}}
    \label{fig:AF}
\end{figure}

Specifically, we focus on the progression of atrial fibrillation (AF) within one month to two years after the initial episode (see Figure \ref{fig:AF}). The early prediction of AF progression in this time-window helps arrhythmia specialists determine whether rhythm control therapies are appropriate for a patient based on their individual risk. In this study, AF progression encompasses all atrial fibrillation subtypes: paroxysmal, persistent, and permanent\footnote{For paroxysmal AF, progression is defined as the recurrence of the arrhythmia, whereas for persistent and permanent AF, it corresponds to the ongoing presence of the condition.}. All methodological steps are evaluated in this context, and the resulting predictive models are compared against traditional clinical scores for AF progression.

To this end, the research is guided by the following core questions:
\begin{itemize}
    \item \textbf{RQ1}: Can the combination of structured EHR data and information extracted from discharge reports enhance and automate the development of ED models?

    \item \textbf{RQ2}: Can free-text discharge reports processed with NLP techniques improve the quality and completeness of structured tabular data derived from EHRs?

    \item \textbf{RQ3}: Can automatically generated silver annotations achieve performance comparable to gold-standard annotations while significantly reducing manual effort?
    
    \item \textbf{RQ4}: Does the proposed methodology produce predictive models that outperform existing clinical scores for AF progression prediction?
\end{itemize}

\section{Related Work}
Disease prediction \cite{sharma2022deep,xie2021multi,yu2023popular,mishra2020chronic} has attracted considerable attention in both the fields of AI and medicine due to its potential to revolutionize ED and personalized medicine \cite{awwalu2015artificial,shah2018next,schork2019artificial,ullah2020applications}. To carry out these studies three steps are necessary: cohort selection, dataset generation and labeling.

For the \textbf{cohort selection} step, recent work has explored automatic methods based on rule driven systems \cite{vydiswaran2019hybrid} as well as NLP approaches that use discriminative language models \cite{soni2021patient} or even generative large language models \cite{cohortGPT}. Our method is related to rule based strategies but incorporates a hybrid module designed to improve generalizability, addressing a common limitation of traditional rule based cohort selection systems \cite{stubbs2019cohort}.

Regarding the \textbf{dataset generation} step, the quality of EHR data is a critical factor in ensuring the reliability and accuracy of predictive models, particularly in the context of medical research. Numerous studies have addressed the common challenges associated with EHR data quality, as well as methods for assessing and improving it \cite{cruz2009data,lewis2023electronic,feder2018data,terry2019basic}. For instance, \cite{jetley2019electronic} suggest leveraging information extraction techniques or manual review of clinical notes to address gaps in structured data. This approach aligns with the hypothesis of the present project, which aims to utilize similar techniques for extracting relevant information from free-text discharge reports to supplement and improve the quality of structured EHR data. 

The detrimental impact of missing values in clinical contexts is widely recognized. Numerous studies emphasize that missing data introduce uncertainty and bias into predictive models, presenting a significant challenge in this field \cite{sterne2009multiple,kahale2020potential,ibrahim2012missing}. Consequently, implementing strategies to reduce the incidence of missing values can effectively mitigate the limitations of current predictive models.

The \textbf{labeling of instances} is a critical component of ED pipelines, yet producing high quality gold standard labels demands extensive expert involvement and manual review. This makes the process time consuming, costly, and difficult to scale. As a result, the use of automatically generated silver standard labels has become increasingly prevalent in ED research. Although these labels are imperfect, prior studies have shown that they can provide sufficient signal to train effective models \cite{wagholikar2020polar,mcdavid2013enhancing}.

In the particular context of \textbf{AF progression prediction}, research in this domain typically focuses on two primary scenarios: the prediction of new-onset AF and the recurrence of AF following therapeutic interventions. AI-driven models have demonstrated remarkable success in predicting incident AF, often outperforming conventional methods \cite{siontis2020will}. These models draw on diverse data sources such as clinical records, cardiac imaging, and electrophysiological data, including the use of EHRs \cite{tseng2021prediction,nadarajah2021predicting,hulme2019development,tiwari2020assessment}. Notably, \cite{sung2022automated} developed a ML model to predict the risk of newly detected AF post-stroke, incorporating both structured variables and unstructured clinical text processed through NLP. 

Most existing studies on AF progression focus on the post-catheter ablation setting. As reviewed by \cite{fan2023predictive}, ML methods have shown strong performance in this context. These approaches often enhance predictive accuracy by combining clinical variables with additional echocardiography inputs \cite{knecht2024machine,zhou2022deep}, features extracted from electrocardiograms (ECGs) \cite{qiu2024deep}, or anatomical data from computed tomography \cite{liu2024use,brahier2023using}.

In addition, despite all the advances in the field of AF progression prediction, a significant gap remains in accurately predicting AF progression following an initial event. This underexplored time window is clinically important, as it can support  first-contact  physicians with patient reference and cardiologists or arrhythmologists in making more informed treatment decisions. Consequently, our study systematically compares traditional clinical scores with our generated models using  automatically obtained datasets.

\section{Resources}
The present research has been possible thanks to the following resources kindly provided by the Basque Public Healthcare System (Osakidetza):

\begin{itemize}
    \item \textbf{Semi-structured discharge reports in Spanish}: $1.2\times10^6$ discharge reports from 2015 to 2020 (see Figure \ref{fig:report}).
    
    \begin{figure}[htb]
        \centering
        \includegraphics[width=5cm]{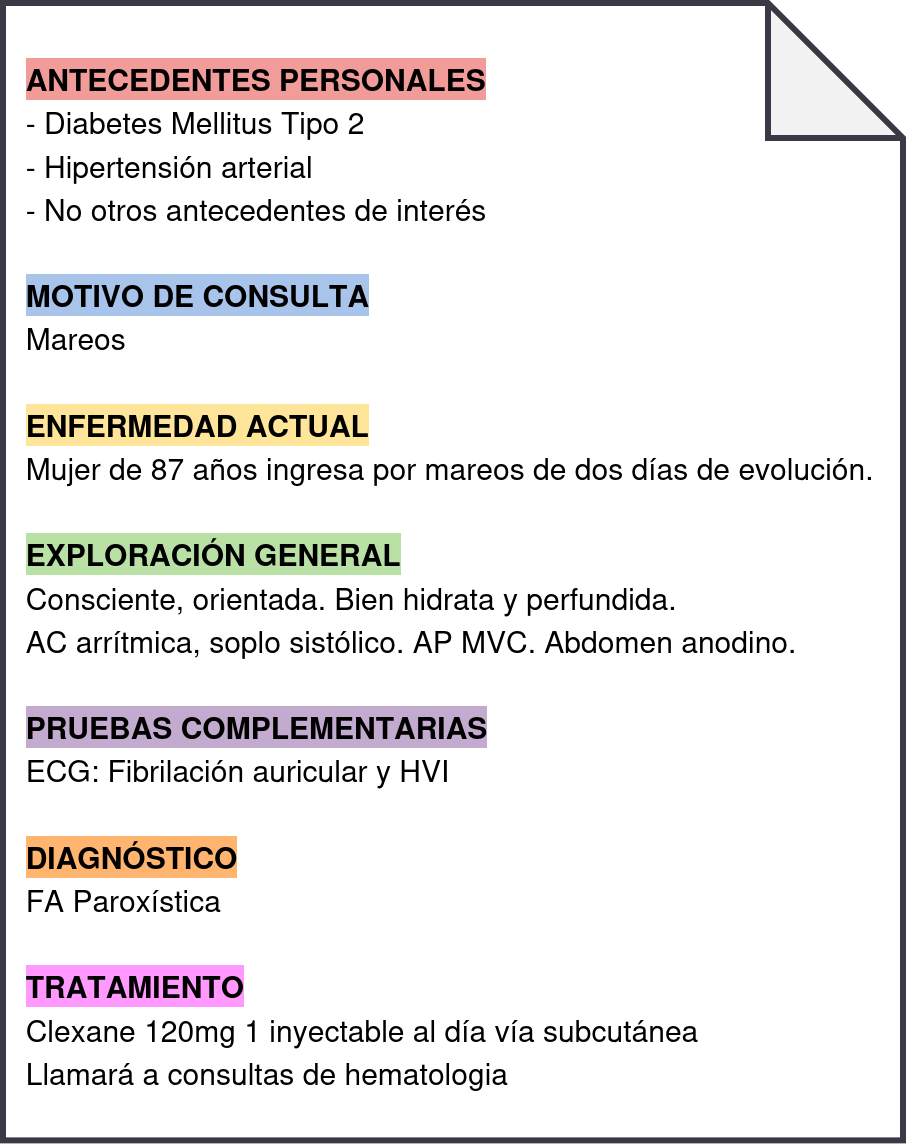}
        \caption{\textbf{Example of a semi-structured discharge report.}}
        \label{fig:report}
    \end{figure}
    
    \item \textbf{Codified structured data}: Clinical data for each patient, encoded by healthcare professionals using standardized coding systems and stored in the Osakidetza Business Intelligence (OBI) platform (see Figure \ref{fig:codified}).

    \begin{figure}[htb]
        \centering
        \includegraphics[width=5cm]{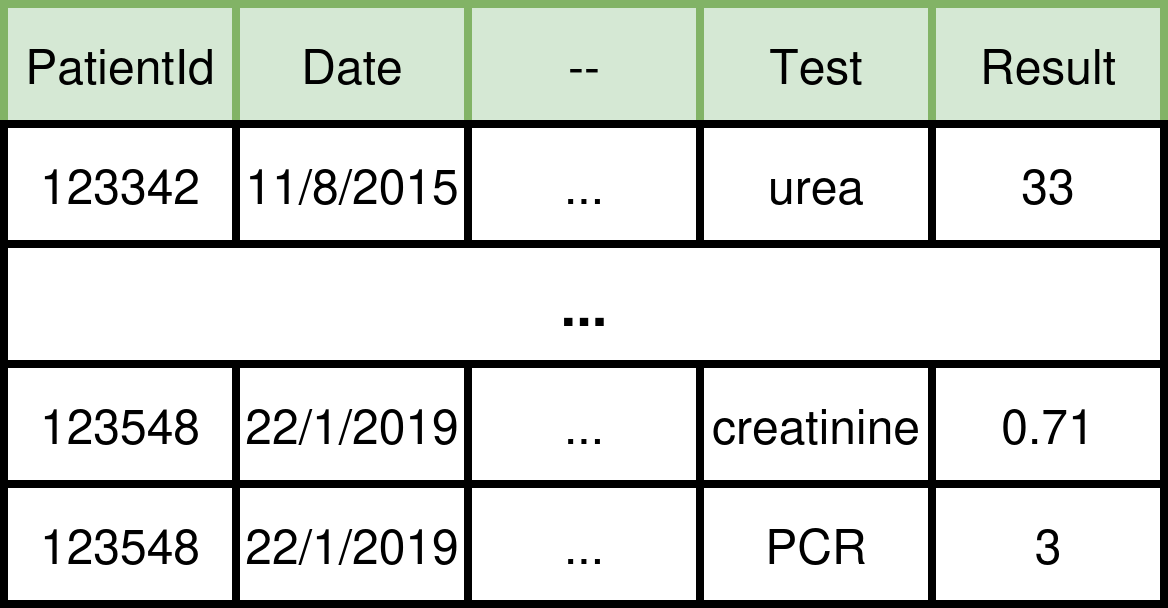}
        \caption{\textbf{Example of the codified structured data.}}
        \label{fig:codified}
    \end{figure}
    
\end{itemize}

\section{Experimental setup}

\begin{figure*}[htb]
    \centering
    \includegraphics[width=0.9\textwidth,clip]{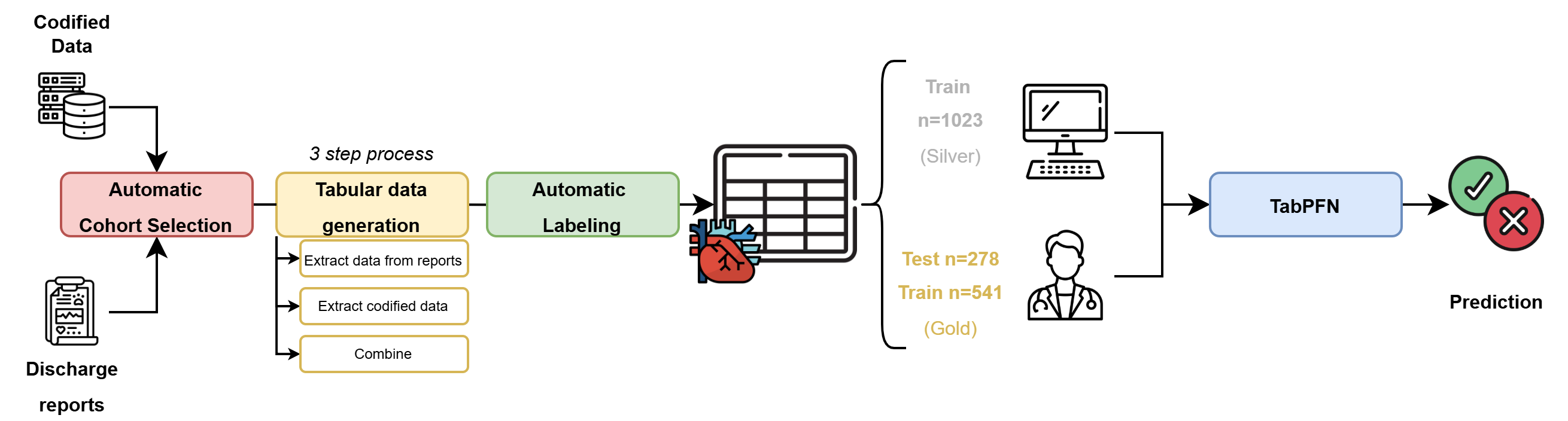}
    \caption{\textbf{End to end overview of the proposed methodology.} The pipeline starts with automatic cohort selection, followed by dataset generation by combining structured EHR data with information extracted from clinical reports. An NLP module performs automatic labeling, and the resulting silver and gold datasets are used to train and evaluate a TabPFN model for AF progression prediction.}
    \label{fig:end-to-end}
\end{figure*}

The experimental setup focuses on the context of AF progression and is divided in three steps, the cohort selection for patients appropriate for the project, the tabular dataset generation step and the automatic labeling step of AF progression (See Figure \ref{fig:end-to-end}).

\subsection{Automatic Cohort selection}

To study AF progression, we identified patients with AF onset, defined as their first documented AF episode with no prior history. Given the errors encountered in \cite{garcia2021role}, we implemented a dual verification approach integrating both structured and unstructured data sources. 

Patients were initially selected via structured EHR data (OBI system), then validated through clinical reports using a two-step NLP approach: first with a fine-tuned EriBERTa encoder-only language model \cite{de2025eriberta}, then a regular expression-based tool for quality control. Both tools are presented in \cite{garcia2025application}.

\subsection{Tabular Dataset Generation and Enrichment}

A total of 85 clinical features were selected for this specific task, encompassing demographic data (7 features), patient history (35 features), laboratory results (18 features), procedures and their outcomes (7 features), treatments (16 features), and AF-related variables, including AF type and the AF progression status, which serves as the target label.

These features include established AF risk factors and general clinical markers aimed at facilitating the prediction of AF progression. While most of them are available and codified within the EHR, several key risk factors, such as left atrial size and even the AF progression status, are not represented in the structured coding system. 

However, some of this information can be found in discharge reports, consequently it is important to retrieve this information to ensure the quality of the predictive models and to reduce the amount of manual annotation needed. 

\begin{figure*}[htb]
    \centering
    \includegraphics[width=0.9\textwidth,clip]{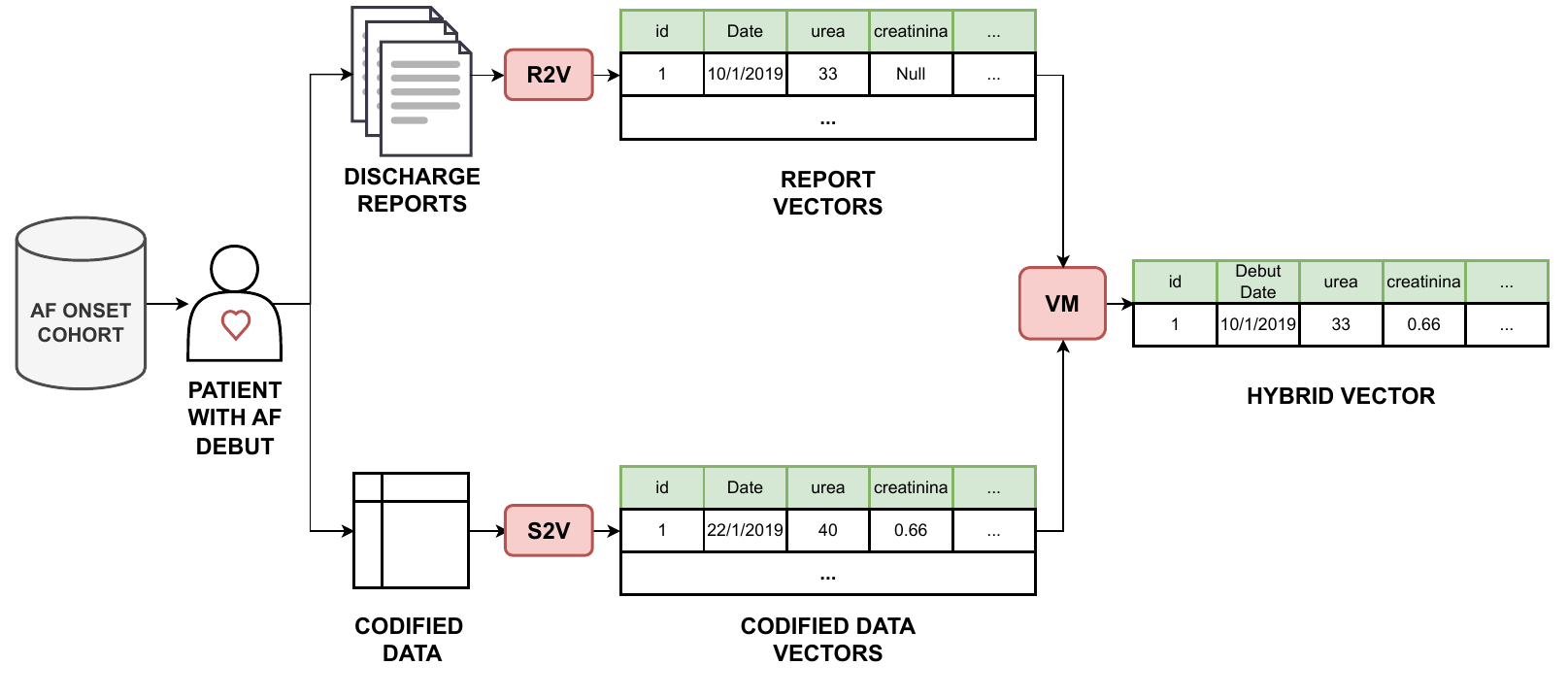}
    \caption{\textbf{Overview of the vector generation process.} For each patient in the AF onset cohort, all discharge reports (free text) and codified data (structured data stored in the Business Intelligence system) are collected and processed using the \textit{Report2Vector (R2V)} and \textit{Structured2Vector (S2V)} tools, respectively. Each tool generates a corresponding set of vectors, which are then merged by the \textit{VectorMerger (VM)} tool to produce a patient-specific vector that integrates both sources of clinical information.}
    \label{fig:IG}
\end{figure*}

The tabular generation pipeline involved three key steps (see Figure \ref{fig:IG}): 
    \begin{enumerate}
        \item \textbf{Extract and structure clinical information from discharge reports using the \textit{Report2Vector} (R2V) pipeline.} This three-step NLP process detailed in \cite{garcia2025application} consists of: (a) section identification, which distinguishes between different parts of the report (e.g., past medical history vs. current episode) \cite{sections2023}; (b) medical entity recognition, which extracts relevant clinical mentions such as symptoms, diagnoses, and procedures, along with negation detection to differentiate between confirmed and ruled-out conditions; and (c) regular expression matching to capture specific patterns of interest. This step returns a table with the 84 predictive clinical variables of interest for AF progression extracted from the discharge reports of the patient.
        \item \textbf{Process structured EHR data from the OBI system using the \textit{Structured2Vector} (S2V) module.} The same 84 predictive features obtained in the previous step are now extracted from the OBI system for each patient and organized into a tabular format.
        \item \textbf{Merge both data sources into unified patient-level vectors using the \textit{VectorMerger}.} The final tabular data contains the codified information from the OBI system enriched with the information extracted from the discharge reports. 
    \end{enumerate}

\subsection{Automatic Labeling}
The automatic labeling process for determining AF progression status consists of the following steps:

\begin{enumerate}
    \item Each patient report is processed following the same approach as the R2V tool (using a section identification module, a medical entity recognition and negation component, and regular expressions). This process extracts the AF status for each consultation date, categorizing it as AF episode, return to sinus rhythm, or no information. As a result, the complete history of the arrhythmia can be reconstructed for each patient.
    \item Using the first documented AF episode as the onset point, subsequent consultations are examined for mentions of new AF episodes or returns to sinus rhythm. AF progression is defined as the occurrence of a new AF episode between one month and two years after the initial diagnosis. Based on this rule, three labels are assigned:
    \begin{itemize}
        \item AF Progression (1): Explicit mention of new AF episode after its onset.
        \item No Progression (0):  Documented sinus rhythm or non-AF ECG findings after the onset.
        \item Excluded (-1): Cases without sufficient evidence to determine progression status.
    \end{itemize}
\end{enumerate}

\subsection{Prediction of AF Progression} \label{sec:pred}
To evaluate the proposed dataset generation methodology, we also apply it to its original purpose: predicting atrial fibrillation (AF) progression within one month to two years after onset.

In this study, we employ a tabular foundation model, also referred to as a Large Tabular Model (LTM) \cite{van2024tabular}. Specifically, we use TabPFN \cite{hollmann2025accurate} due to its ability to efficiently handle complex, small-scale tabular datasets while leveraging prior knowledge through pretraining.

During preliminary experimentation, several ML architectures were evaluated, including Support Vector Machines, Random Forests, and XGBoost. However, due to the high dimensionality of the predictive features and the substantial proportion of missing values, these models did not yield satisfactory results and were subsequently discarded. The experiments also explored various missing value imputation strategies (such as mean and median imputation, as well as logistic regression–based imputation) none of which outperformed the TabPFN performance. Additionally, data preprocessing techniques including feature standardization, feature selection, and different sampling strategies (undersampling, oversampling, SMOTE, and Tomek). Moreover, TabPFN incorporates internal data preprocessing strategies (including handling missing values and scale normalization) as well as an integrated hyperparameter optimization routine, which we utilized.

We evaluate performance differences between two datasets: the original dataset derived from codified EHR information, and the enriched dataset that incorporates features extracted from discharge reports.

In addition, we compare model performance when using automatically generated silver-standard annotations with that obtained using gold-standard labels manually annotated by a cardiologist. The characteristics of the dataset used in these experiments are summarized in Table \ref{tab:dataset}.

\begin{table}[htb]
    \centering
    \begin{tabular}{l r r }
        \hline
        & \textbf{Size} & \textbf{\% Positives}\\ \hline
        \textit{Train-Silver} & 1023 & 65.40\% \\
        \textit{Train-Gold} & 541 & 66.17\%\\
        \textit{Test} & 278 & 64.03\% \\ \hline
    \end{tabular}
    \caption{\textbf{Characteristics of the dataset.} The second column indicates the number of patients and the last column the percentage of AF progression.}
    \label{tab:dataset}    
\end{table}

\subsubsection{Evaluation}

Model performance was assessed using both accuracy and the Matthews Correlation Coefficient (MCC). While accuracy measures the proportion of correct predictions, it can be misleading in datasets with imbalanced classes. MCC provides a more robust evaluation by incorporating all elements of the confusion matrix (true positives, true negatives, false positives, and false negatives) yielding values between –1 (complete disagreement) and 1 (perfect agreement).

In medical prediction tasks, MCC is especially valuable because it evaluates the quality of predictions for both classes simultaneously. Unlike the F1-score, which focuses solely on the positive class, MCC captures the balance between correctly identifying patients with and without the condition. This distinction is crucial in clinical contexts, where recognizing true negatives is as important as detecting true positives to prevent unnecessary interventions, costs, and patient distress. Consequently, MCC offers a more comprehensive and reliable measure of model performance in healthcare applications.

Moreover, to evaluate the clinical relevance of our predictive models, we compared them with established clinical scores for AF progression: CHADS2-VASc\footnote{Used for stroke risk stratification in AF patients.} \cite{lip2010refining}, HATCH\footnote{Used for AF onset prediction.} \cite{de2010progression}, and APPLE\footnote{Used for AF recurrence risk prediction after catheter ablation.} \cite{kornej2015apple}.

As these scores produce numerical values rather than direct classifications, a threshold of $\geq$ 2 was applied to convert them into binary outcomes. The scores were calculated using the generated tabular data and the formulas from their original publications.

\section{Results and Discussion}
In this section, we present the results of the proposed methodology, analyzing its impact on the proportion of missing values in the final dataset as well as the differences between the automatic and gold-standard annotations. We also report the outcomes of the AF progression prediction experiments performed using the different versions of the dataset.

\subsection{Dataset Enrichment} \label{sec:disc_1}
\begin{figure*}[htb]
    \centering
    \includegraphics[width=0.98\textwidth,clip]{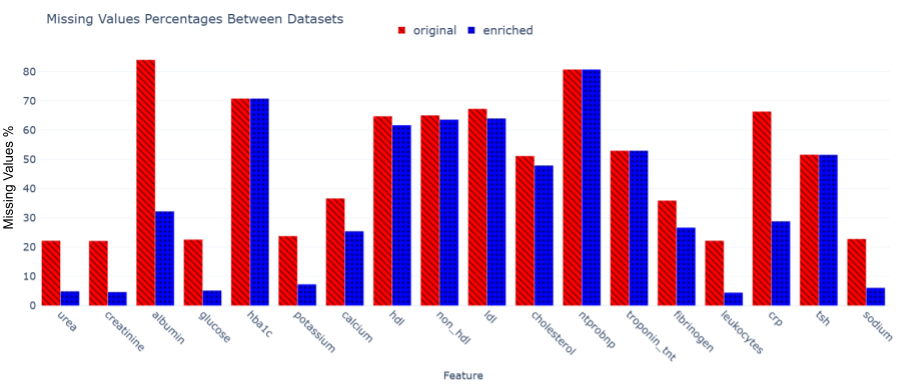}
    \caption{\textbf{Amount of missing values. Difference in percentage between the \textcolor{blue}{original} and \textcolor{red}{enriched} datasets.} The bar plots illustrate the recovery of features that were absent in the original codified dataset but retrieved from the information contained in the discharge reports.}
    \label{fig:enriched}
\end{figure*}

The enriched dataset using the information of discharge reports improves considerably the amount of missing values and general recall of variables (see Figure \ref{fig:enriched}). The comparison between the original and enriched datasets highlights the significant impact of incorporating information extracted from discharge reports. 

As shown in the figure \ref{fig:enriched}, the proportion of missing values in the original dataset (blue bars) is notably reduced in the enriched dataset (red bars) across most laboratory variables, indicating a substantial improvement in data completeness. Variables such as albumin, CRP, and NT-proBNP (which originally presented high levels of missingness) show a marked increase in data availability after enrichment. This suggests that the integration of textual information helps recover clinically relevant details often absent from coded records.

%Similarly, the second and third figures (see Figures \ref{fig:history} and \ref{fig:treatments}) demonstrate a clear increase in the proportion of positive cases for categorical variables regarding past medical history and treatments. These findings reinforce the value of unstructured text as a complementary data source, capable of capturing patient conditions frequently underrepresented in structured codified EHR fields. 

The recovery of past medical history features increased substantially, rising from an average of 2.62\% positives in the original dataset to 14.25\% positives in the enriched version (an absolute gain of 11.63 percentage points). Among all feature categories, treatment-related variables benefited the most from the enrichment process. Their mean recall rose from 1.73\% in the original dataset to 45.05\% in the enriched dataset.

Furthermore, certain clinically relevant variables with high predictive value were completely missing from the original codified dataset, as they are not recorded within the OBI system. For example, the left atrial size is a demonstrated predictor for AF recurrence and through the use of NLP extraction from discharge reports, this feature achieved a recall of 41.8\%, which is remarkable given that not all patients have this information documented in their discharge summaries.

Missing values in EHRs can arise from multiple sources beyond human error. They may result from data integration issues across systems, variations in clinical documentation practices, or patient-related factors such as refusal or missed tests. System design limitations, such as non-mandatory fields also might contribute, as do temporal gaps when data is pending or historical records are unavailable. Overall, the enrichment process not only enhances data completeness but also improves the representativeness of clinical variables, providing a stronger foundation for downstream predictive modeling.

\subsection{Automatic Labeling} \label{sec:disc_2}
The automatic labeling approach for patients’ AF progression status demonstrated an accuracy of 0.82 relative to the manually annotated test set. Achieving an accuracy of 0.82 is notable, especially considering that the manual annotations incorporate the full spectrum of patient information, including electrocardiograms and other details that may not be fully documented in the discharge reports. This means our methodology is able to capture and label patient AF progression status accurately despite the possible incompleteness of the discharge data.

\subsection{AF Progression Prediction} \label{sec:disc_3}

To further assess the utility of our methodology and its application in future studies of early prediction of diseases we performed three experiments for AF progression prediction.  The three experiments involve different versions of the generated dataset to evaluate how the proposed methodology impacts the results of AF progression. The first experiment uses the original dataset (1023 instances) extracted solely from codified EHRs with the silver annotation. The second experiment uses the enriched dataset (1023 instances), which combines codified information and discharge report data along with our silver-automatic annotation. Finally, the third experiment uses a smaller subset of the enriched dataset (541 instances) that has been manually annotated with gold-standard labels by a cardiologist.

As explained in section \ref{sec:pred}, we use the TabPFN architecture for the three experiments. The results are available in Table \ref{tab:results}.

\begin{table}[htb]
    \centering
    \begin{tabular}{l r r }
        \hline
        \textbf{Dataset} & \textbf{Accuracy} & \textbf{MCC}\\ \hline
        \textit{Original-Silver} & 0.65 & 0.11 \\
        \textit{Enriched-Silver} & 0.66 & 0.20\\
        \textit{Enriched-Gold} & 0.66 & 0.21 \\
        %\textit{Enriched-Hybrid} & 0. & 0. \\
        \hline
    \end{tabular}
    \caption{\textbf{Results obtained in the AF progression experiments.}}
    \label{tab:results}    
\end{table}

Using the original dataset with silver annotations derived solely from codified EHRs, the model achieved a moderate accuracy of 0.65 and a low MCC of 0.11, indicating limited predictive power when relying exclusively on structured EHR data. Incorporating discharge report information in the enriched dataset with silver-automatic annotations led to an improvement in accuracy (0.66) and doubled the MCC (0.20), suggesting that integrating unstructured clinical text adds relevant information. In medical prediction tasks, this increase is particularly meaningful, as MCC reflects the overall reliability of the model across all possible outcomes. Doubling the MCC indicates a substantial improvement in the model’s ability to correctly identify both patients at risk and those not at risk, which is crucial in clinical settings where misclassification can lead to missed diagnoses or unnecessary treatments.

Finally, using the manually annotated gold-standard subset resulted in a similar accuracy of 0.66, with a slight increase in MCC to 0.21. Although this subset contains almost half the number of instances, the higher-quality annotations improve the correlation between predictions and true labels. However, manual annotation is time-consuming and requires expert knowledge. Therefore, an automatic method that achieves comparable performance, even if it requires more instances, remains advantageous.

The predictive models consistently outperform the traditional clinical scores in both accuracy and MCC (see Table \ref{tab:clinical_scores}). Both CHADS2-VASc and HATCH achieved an accuracy of 0.60 with very low MCC values (–0.0052 and 0.0832, respectively). APPLE performed worse, with an accuracy of 0.48 and an MCC of 0.05.

\begin{table}[htb]
    \centering
    \renewcommand{\arraystretch}{1.2}
    \begin{tabular}{lccc}
        \hline
        & \textbf{ACC} & \textbf{MCC}\\
        \hline
        \textit{CHADS2-VASc} & 0.6043 & -0.0052 \\
        \textit{HATCH}& 0.6043 & 0.0832\\
        \textit{APPLE} & 0.4820 & 0.0510 \\ \hline
    \end{tabular}
    \caption{\textbf{Results obtained in the AF progression experiments by the clinical scores.}}
    \label{tab:clinical_scores}
\end{table}

These results highlight that data-driven predictive models, even with automatic annotations, can capture patterns in AF progression that traditional clinical scores fail to detect. While clinical scores are useful for quick risk stratification, their limited consideration of patient-specific information and simplified scoring rules result in lower predictive performance. In contrast, integrating structured EHR data with textual discharge reports allows the models to leverage a richer set of features, yielding more reliable predictions, particularly for imbalanced outcomes like AF progression. This suggests that such models could serve as valuable decision-support tools, complementing or enhancing existing clinical scores.

\section{Conclusions}

This study provides evidence that discharge reports, when processed with NLP techniques, can play a significant role in supporting early prediction tasks and enhancing clinical data quality. Addressing each research question:

\textbf{RQ1: Can the combination of structured EHR data and information extracted from discharge reports enhance and automate the development of early disease prediction models?} The study demonstrates that integrating structured EHR data with NLP extracted information from discharge reports both facilitates automation of the data preparation pipeline and yields measurable improvements in model performance and data completeness. The proposed pipeline automates cohort selection, tabular feature generation and outcome labeling by combining codified EHR vectors with features extracted from free text. Practically, this automation reduces the need for manual case review and manual label assignment, streamlining the end to end process required to build early prediction models. In the AF progression experiments the enriched dataset that merged codified data with discharge report features produced a higher MCC than the codified dataset alone (MCC 0.20 versus 0.11) while accuracy rose slightly from 0.65 to 0.66. Automatic labeling via the pipeline also produced a high agreement with manual annotation, with an automatic labeling accuracy of 0.82 on the manually annotated test set. These results indicate that the integrated approach both automates previously manual steps and supplies additional predictive signal for the model.

\textbf{RQ2: Can free-text discharge reports processed with NLP techniques improve the quality and completeness of structured tabular data derived from EHRs} The integration of NLP-extracted information with existing tabular data substantially reduces missingness and increases recall for many clinically relevant variables that are underrepresented in codified fields. The article reports a clear reduction in the proportion of missing values for many laboratory and categorical features after enrichment with discharge report information. Laboratory variables that originally had high missingness, such as albumin, C reactive protein and NT proBNP, showed marked increases in availability. Similarly, recall for past medical history and treatment features rose noticeably in the enriched dataset. The VectorMerger approach produces unified patient vectors where features absent from the codified system are recovered from text, improving completeness and representativeness. This adresses the gaps caused by system limitations, documentation variability, or patient-related factors, the enrichment process enhances data completeness and representativeness, providing a more reliable foundation for predictive modelling.

\textbf{RQ3: Can automatically generated silver annotations achieve performance comparable to gold-standard annotations while significantly reducing manual effort?} Automatically generated silver annotations achieve performance close to gold-standard annotations in model evaluation and greatly reduce manual effort. The similarity of the enriched-silver and enriched-gold results, especially in MCC, indicates that silver annotations can approach gold performance in this problem setting. Given comparable model performance and that manual gold labeling is costly and time consuming, silver labeling permits larger training sets and faster iteration.

\textbf{RQ4: Does the proposed methodology produce predictive models that outperform existing clinical scores for AF progression prediction?} Models trained with the enriched dataset outperform traditional clinical scores for AF progression on both accuracy and MCC. Across the experiments, data driven models consistently outperformed the clinical scores considered. The TabPFN model trained on the enriched dataset achieved accuracy 0.66 and MCC 0.20. By contrast, CHADS2 VASc and HATCH each had accuracy approximately 0.60 with very low or near zero MCC values. The APPLE score performed worse with accuracy 0.48. These results indicate that ML models capture predictive patterns that simplified clinical scores do not.

Overall, this study highlights the potential of combining structured EHR data with NLP-extracted information from discharge reports to enhance early prediction tasks. Such an approach not only supports more robust and comprehensive early prediction models for AF progression but also lays the groundwork for applying similar methodologies to other diseases, ultimately contributing to more informed and data-driven clinical decision-making.

\section{Future Work}
The current study was limited to AF progression; applying the same pipeline to other chronic and acute conditions will be essential to assess its generalizability and robustness. In addition, external validation using datasets from different hospitals and healthcare systems will help evaluate the methodology under diverse documentation practices and data standards. Expanding the study to multiple institutions and involving a larger number of expert annotators would further strengthen the robustness and external validity of the proposed framework. However, generating high quality expert labels is a time consuming and resource intensive process, which was beyond the scope of the present work. Therefore, broader expert involvement and multi center validation are important directions for future research.

Future work will also explore the incorporation of advanced NLP approaches, including large language models (LLMs), into the existing pipeline. The rapid evolution of these architectures offers promising opportunities to further enhance the accuracy of feature extraction and automatic labeling.

In the specific context of AF progression prediction, future research could investigate multimodal approaches that process structured tabular data and unstructured clinical text jointly when estimating patient risk. Moreover, integrating temporal information from longitudinal patient histories may improve the modeling of disease dynamics and lead to more precise and clinically meaningful predictions.

\section*{Acknowledgements}
This work has been partially supported by the HiTZ Center and the Basque Government, Spain (Research group funding IT1570-22) as well as by MCIN/AEI/10.13039/5011 00011033 Spanish Ministry of Universities, Science and Innovation  by means of the projects:
EDHIA PID2022-136522OB-C22 (also supported by FEDER, UE).

A. G. Domingo-Aldama has been funded by the Predoctoral  Training Program for Non-PhD Research Personnel grant of the Basque Government (PRE\_2024\_1\_0224).

A. García Olea has been funded by BioBizkaia grant under the code BB/I/PMIR/24/001

\clearpage
\bibliographystyle{fullname}
\bibliography{bibilography}

\appendix

\end{document}